\documentclass{isprs} 
\usepackage{subfig}
\usepackage{setspace}
\usepackage{geometry} 
\usepackage{epstopdf}
\usepackage{url}
\usepackage{amsmath}
\usepackage{multirow}
\usepackage{color}
\usepackage[labelsep=period]{caption}  
\usepackage[british]{babel} 
\usepackage[hang]{footmisc}
\usepackage{natbib}

\begin{document}

\title{Image-based Deep Learning for the time-dependent prediction of fresh concrete properties}
\date{}


\author{Max Meyer$^1$, Amadeus Langer$^1$, Max Mehltretter$^1$, Dries Beyer$^2$, Max Coenen$^2$, Tobias Schack$^2$, Michael Haist$^2$, Christian Heipke$^1$}

\address{$^1$Institute of Photogrammetry and GeoInformation, Leibniz University Hannover, Germany\\
(meyer, langer, mehltretter, heipke)@ipi.uni-hannover.de\\
$^2$Institute of Building Materials Science, Leibniz University Hannover, Germany\\
(d.beyer, m.coenen, t.schack, haist)@baustoff.uni-hannover.de}




\abstract{Increasing the degree of digitisation and automation in the concrete production process can play a crucial role in reducing the CO$_2$ emissions that are associated with the production of concrete. In this paper, a method is presented that makes it possible to predict the properties of fresh concrete during the mixing process based on stereoscopic image sequences of the concretes flow behaviour. A Convolutional Neural Network (CNN) is used for the prediction, which receives the images supported by information on the mix design as input. In addition, the network receives temporal information in the form of the time difference between the time at which the images are taken and the time at which the reference values of the concretes are carried out. With this temporal information, the network implicitly learns the time-dependent behaviour of the concretes properties. The network predicts the slump flow diameter, the yield stress and the plastic viscosity. The time-dependent prediction potentially opens up the pathway to determine the temporal development of the fresh concrete properties already during mixing. This provides a huge advantage for the concrete industry. As a result, countermeasures can be taken in a timely manner. It is shown that an approach based on depth and optical flow images, supported by information of the mix design, achieves the best results.}

\keywords{Fresh concrete properties, Building materials, Stereoscopy, Deep learning, Time dependency.}

\maketitle


\section{Introduction}\label{MANUSCRIPT}
Reducing CO$_2$ emissions poses a major challenge for the construction industry. Concrete production in particular plays a decisive role here. Concrete is one of the most widely used building materials in the world, and the production of its constituent cement is in itself responsible for approx. 6.7 \% of global anthropogenic CO$_2$ emissions \citep{IEAcement}. Many approaches are therefore focusing on reducing the cement content by using substitute materials. As a consequence there is an increasing trend for concrete to no longer consist of just three materials (cement, aggregate and water), as was originally the case, but of several additional materials in order to reduce the amount of cement needed. In turn, this also leads to increasingly complex mix designs, which causes a potentially less robust concrete \citep{Gonzalez2018}. Consequently, the control of concrete properties becomes more difficult, especially the fresh concrete properties.\\ \\
Existing quality assurance measurements are not well suited to overcome this new challenge. These measurements are usually carried out manually and only for batch samples, both of which increases the uncertainty associated to the measured properties. Since the properties during concrete placement are decisive for the quality of the resulting building component, the measurements are carried out directly before placement on the construction site, where there is no longer any possibility of significantly adjusting the fresh concrete properties. Deviations of the actual concrete quality from the target properties may lead to a rejection of the batch, resulting in an inefficient production which wastes a lot of resources. In the opinion of the authors, a key change to increase the sustainability of concrete production is to digitize and automate the production process including quality control \citep{Haist2022}. Since in comparison to other industries, the construction industry is one of the least digitized and automated industries in the world
\citep{Green2016}, we see huge potential for improvement, however requiring sensor based insight into the material properties.\\
\\
The ReCyCONtrol\footnote{\url{https://www.recycontrol.uni-hannover.de/en/}} research project addresses this lack of digitization and automation in the concrete sector. 
One part of the project focusses on the prediction of the fresh concrete properties. Since the moment of production, i.e. during the mixing process, offers the most opportunities of adjusting the concrete properties in case of quality deviations, the prediction of these properties should be done during the mixing process. Also, as the properties of the concrete may further change between the mixing process and its placement, due to the cements chemical hydration process, the behaviour of the properties after mixing must be modeled over time. We therefore formulate the goal of predicting the future properties of the concrete, e.g.\ for the time of placement, already during the production step. If deviations to the target properties at the time of placement are estimated in this way, countermeasures in the form of chemical additives can be used to change the properties to reach the desired values.\\
\\
To reach this goal of predicting the fresh concrete properties at the time of placement during the mixing process, we use optical sensors coupled with a corresponding photogrammetric processing chain: We use a stereo camera system to observe the concretes flow behaviour during the mixing process. The stereo images then serve as input for a deep learning method, which predicts the properties of the concrete as a function of time. To describe the properties, we use one parameter for the consistency and two for the flow behaviour. The slump flow diameter $\delta$, which is measured by the slump test \citep{flowtabletest}, represents the consistency of the concrete. The flow behaviour of the concrete, e.g.\ during the mixing process, is defined by its rheological parameters. In more detail, concrete, as a non-Newtonian fluid, can be described by the Bingham fluid with the rheological parameters yield stress $\tau_0$ and plastic viscosity $\mu$ \citep{yahia2016concrete}. The values of these parameters can be derived from the flow curve, which can be measured in batch experiments by a rheometer. The yield stress is a measure of how much stress has to be applied to set a liquid in motion. The plastic viscosity describes the viscosity of a liquid at high shear rate. The slump flow diameter depends on the rheological parameters. In particular the yield stress and the slump flow diameter are assumed to be correlated \citep{wallevik2006relationship}.\\
\\
In this paper a novel method for the prediction of the slump flow diameter, the yield stress and the plastic viscosity is proposed. The prediction is carried out by a convolutional neural network (CNN), which receives stereo camera observations of the concretes flow at a certain sample age as input. Furthermore, we add temporal information and information from the mix design to the input. For the prediction of the temporal evolvement of the fresh concrete properties, the temporal information is crucial. Using this information, the CNN learns implicitly to model the time-dependent behaviour of the properties. This enables the possibility of predicting the fresh concrete properties for arbitrary points in time,  e.g.\ the time of placement.\\
\\
The paper is structured as follows: We first give an overview of the current research in the field of digitizing and automating the concrete sector. Our methodology is then described in section \ref{methodology}. The data set and how it is generated are presented in section \ref{data}. Section \ref{experiments} shows the experiments and contains a discussion of the results. Section \ref{conclusion} concludes the paper and gives an outlook for possible future work.
 
\sloppy
\section{Related work}\label{relwork} 
In recent years, the automation of concrete quality assurance has received increasing attention. In \cite{song2020deep}, image segmentation is used to determine properties of the hardened concrete. \cite{Coenen2021a} used image segmentation to determine the particle size of the aggregates in the hardened concrete. Although these methods have the potential to automate the current quality assurance measurements, at this stage no countermeasures can be applied if deviations to the target properties are detected.\\
\\
To ensure the quality of the fresh concrete, traditional quality assurance measurements like the slump test \citep{flowtabletest} and rheometer measurements are typically employed. However, these methods are labor intensive and are associated with relatively high uncertainties. In \cite{tuan2021situ} a method is proposed to automate the slump test: Instead of measuring the diameter of the spread concrete manually, a stereo camera system records images of the spreading concrete and the diameter is determined using image processing. 
The authors argue that replacing the manual measurement with an imaging system improves the accuracy of the result and reduces the workload. \cite{yoon2023framework} propose a method for analysing cement paste with a similar set up. Instead of a stereo camera, a depth camera is used to record a point cloud of the cement paste after the slump test. The point cloud is used to extract the diameter, spread height and curvature. These parameters are used as input for a deep learning algorithm to predict yield stress, plastic viscosity, adsorption ratio of superplasticizer and bleeding. 
\cite{schack2023bildbasierte1, schack2023bildbasierte2, schack2023bildbasierte3} take this work one step further and use images taken from the spread flow of the concrete and are not only able to derive the spread flow diameter of the concrete but also information on the concrete composition. Therefore the surface roughness of the spread flow cake is analysed in order to derive e.g.\ the content of the coarse particle contained in the concrete. In \cite{coenen2024deep}, a method is presented, in which a camera observes the channel flow of the fresh concrete at the outlet of a mixing truck. Spatio-temporal flow fields are generated from the recorded images, which contain information about the flow behaviour of the concrete. A CNN predicts the fresh concrete properties on the basis of the spatio-temporal flow fields.
The disadvantage of these methods is, that they are applied post-production, meaning that the concrete still has to be discharged and new concrete has to be produced, if significant deviations to the target properties are detected. \\
\\
To overcome this drawback, the properties of the fresh concrete have to be predicted before or during the mixing process. There are two main procedures to achieve this goal. One approach is to perform the prediction on the basis of the mix design information of the concrete. The concrete mix design contains the exact content (in kilograms) of the individual materials used to produce the concrete. The type and concentration of the materials have a major impact on the properties of the concrete. \cite{chidiac2009plastic} summarize the most common models to determine the plastic viscosity based on the mix design. It is shown that the results vary between different models. The most recent methods based on the mix design use machine learning, and in particular deep learning algorithms. Methods like least squares support vector machines (LSSVM) and particle swarm optimization (PSO) \citep{nguyen2020prediction}, extreme learning machines \citep{kina2021comparison}, random forests and XGBoost \citep{zhang2022predicting,hosseinzadeh2023prediction} as well as multi layer perceptrons (MLP) \citep{navarrete2023predicting} are used for this purpose. In \cite{nguyen2020prediction} and \cite{navarrete2023predicting} the input information from the mix design is extended with temporal information, representing the time difference between the mixing process and the time at which the properties are to be determined, to take into account the change of properties over time. The mix design contains valuable information for the time-dependent change of the properties (e.g.\ the additive content). Although these methods achieve promising results, they omit essential information as e.g.\ possible variations in the properties of the employed constituents, even though there has been progress in this field in recent years, e.g.\ \cite{Coenen2023a,lux2023classification}.\\
\\
The second main procedure is to predict the fresh concrete properties based on images of the fresh concrete acquired during the mixing process. \cite{li2014method} showed that it is possible to estimate the slump flow and the V-funnel flow time from images recorded during the mixing process by using classical image analysis methods, namely frame difference and watershed segmentation. \cite{ding2018deep} show that also deep learning methods, here a combination of a CNN and a long short-term memory network (LSTM) based on image sequences, are applicable. \cite{yang2021estimating} and \cite{guo2022real} employ another combination of CNN and LSTM with image sequences to predict the slump value and slump flow value respectively the plastic viscosity, while \citep{gao2023numerical} use semantic segmentation in combination with a residual neural network for single images for the prediction of the slump class. In \cite{ponick2022image}, a stereo camera set up is used to observe the mixing process of ultra sonic gel, a often employed surrogate for concrete. The stereo camera observations are used as input for a CNN. The results show that the 3D information derived from the stereo images is valuable for the prediction of the flow curves. Although, these methods show promising results and are not prone to uncertainties in the mix design, they do not take into account the time dependency of the fresh concrete properties.\\
\\
To the best of the authors' knowledge, there is no method to date that uses both images and information from the mix design to perform a time-dependent prediction of fresh concrete properties.
\sloppy
\section{Deep Learning for fresh concrete properties prediction}\label{methodology} 
\subsection{Overview}

We aim to predict fresh concrete properties based on image observations. 
Stereo cameras record synchronized RGB image pairs of the concretes flow during the mixing process. Each image pair is used to generate an orthophoto $O$ and a depth map in the form of a digital elevation model (DEM) $D$, which contains 3D information about the surface of the fresh concrete. The photogrammetric process to generate $O$ and $D$ is carried out using a commercial software, namely \textit{Agisoft Metashape}\footnote{\url{https://www.agisoft.com/}}. To reduce the computational time during training, $O$ is transformed into a panchromatic image. $O$ and $D$ are extended by an optical flow image $OF$, which contains the displacement of each pixel between the current and the subsequent orthophoto. This allows to additionally include explicit motion information representing the flow behaviour of the concrete into the network. To generate the optical flow images $OF$, for each $O_i$, where $i$ represents the time step, the optical flow $OF_{i,i+1}$ between $O_i$ and $O_{i+1}$ is computed. For this purpose the method presented in \citep{farneback2003two} is used. The input images are stacked to obtain the input set $[O_i, D_i, OF_{i,i+1}]$. The input set is treated as an image with four channels, one channel each for $O_i$ and $D_i$, and two channels for the optical flow image $OF_{i,i+1}$ (one channel each for the displacements of the pixels in the x and y direction, respectively). 
\\
\\
For the purpose of modeling the time-dependent behaviour of the fresh concrete properties, temporal information $\Delta_t$ is introduced. In this context, $\Delta_t$ represents the time difference between the age of the sample at which the image pair for generating $O_i$ and $D_i$ is acquired and the age for which the fresh concrete properties are to be predicted. Besides the temporal information $\Delta_t$, information from the mix design $m$ is added as additional input. $m$ contains information about the water-cement (mass) ratio, the grading curve of the aggregate particles, the paste content, the admixture content and the time difference between starting the mixing process and the image acquisition. To take into consideration the influence of different mixing velocities (i.e. the speed of the mixing tools) and frame rates of the imaging sensors, these parameters are both added to $m$. For numerical reasons, the inputs and the reference values are normalized. Consequently, the outputs of the CNN are the predicted normalized values for the slump flow diameter $\delta_{\Delta_t}$ and the rheological parameters yield stress $\tau_{0,\Delta_t}$ as well as plastic viscosity $\mu_{\Delta_t}$ at the age of the sample defined by $\Delta_t$. These parameters are summarised in the target state vector $C=[\delta_{\Delta_t},\tau_{0,\Delta_t},\mu_{\Delta_t}]$. 

\subsection{Network architecture}\label{sec:Network architecture}
\begin{figure}[h!]
\centering
\includegraphics[width=1.0\columnwidth]{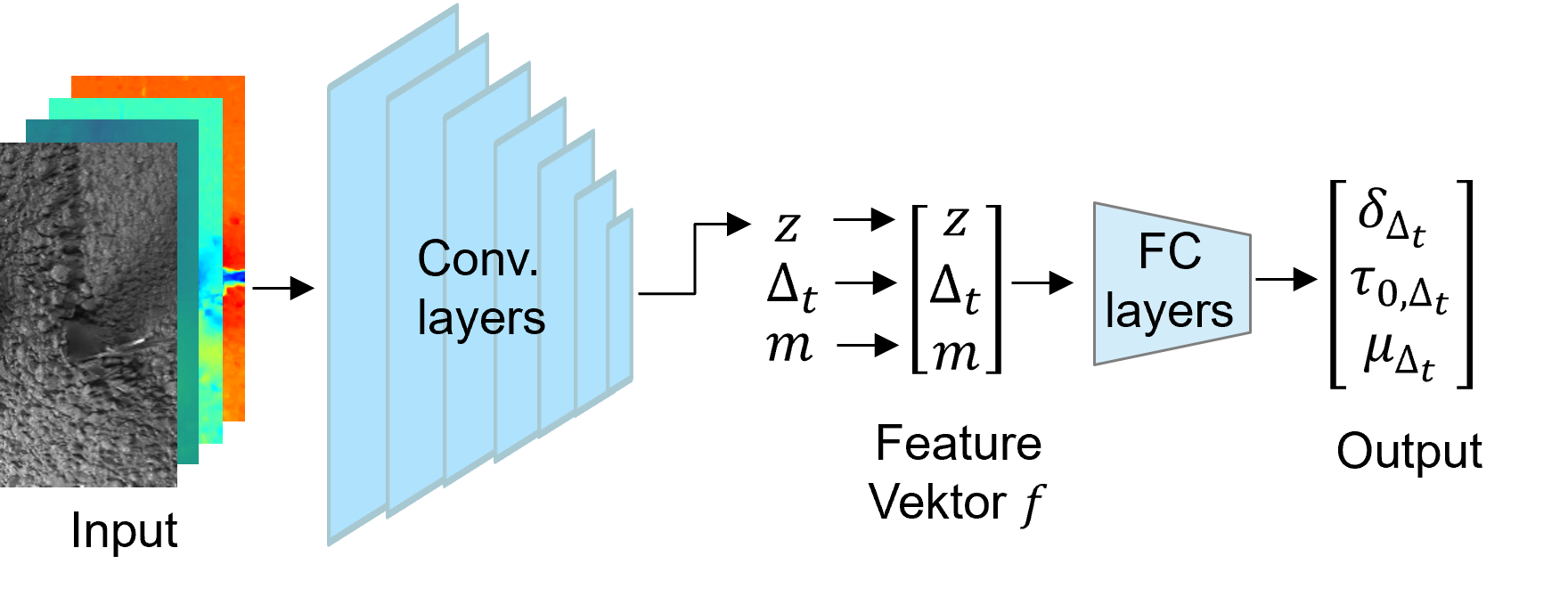}
\caption{CNN architecture of developed method.}
\label{fig:architectureCNN} 
\end{figure}

We make use of a CNN for the prediction of the state vector C, consisting of seven convolutional layers which are followed by three fully connected layers (FC layers). The architecture of the CNN is based on the CNN presented in \cite{ponick2022image}. As the problem is less complex than usual classification and segmentation tasks and we only have a relatively small amount of training data, we limit the number of unknowns by using comparatively few layers. The results in \cite{ponick2022image} support this approach. A high-level overview of the architecture is shown in Fig.~\ref{fig:architectureCNN}. The convolutional layers have a kernel size of 5x5 and a stride of 2 each, and are followed by batch normalisation and a Rectified Linear Unit (ReLU) activation function. The number of neurons in the FC layers decreases linearly from 660 down to the three output neurons in the output layer. Each FC layer has a leaky ReLU activation function using a slope of 0.2. No batch normalization is used between the FC layers.\\
\\
The input set $[O_i, D_i, OF_{i,i+1}]$ is fed to the convolutional layers. The layers extract features to produce the flattened feature embedding $z$ with a length of 640 elements. $\Delta_t$ and $m$ are added in a late-fusion manner to $z$. By concatenating $z$, $\Delta_t$ and $m$, we obtain the feature vector $f$ which is passed to the FC layers as input. The FC layers map $f$ to the time-dependent output parameters $\delta_{\Delta_t}$,$\tau_{0,\Delta_t}$, and $\mu_{\Delta_t}$ of the target state vector $C$. This approach was chosen because the FC layers form a MLP and MLPs are suitable for a time-dependent prediction of the static yield stress of cement paste based on the mix design information, temporal information and information on properties of the raw materials \citep{navarrete2023predicting}.
\begin{figure*}[h]
\centering
\subfloat[Schematic overview of the experimental set up.] {\includegraphics[width=0.5\textwidth]{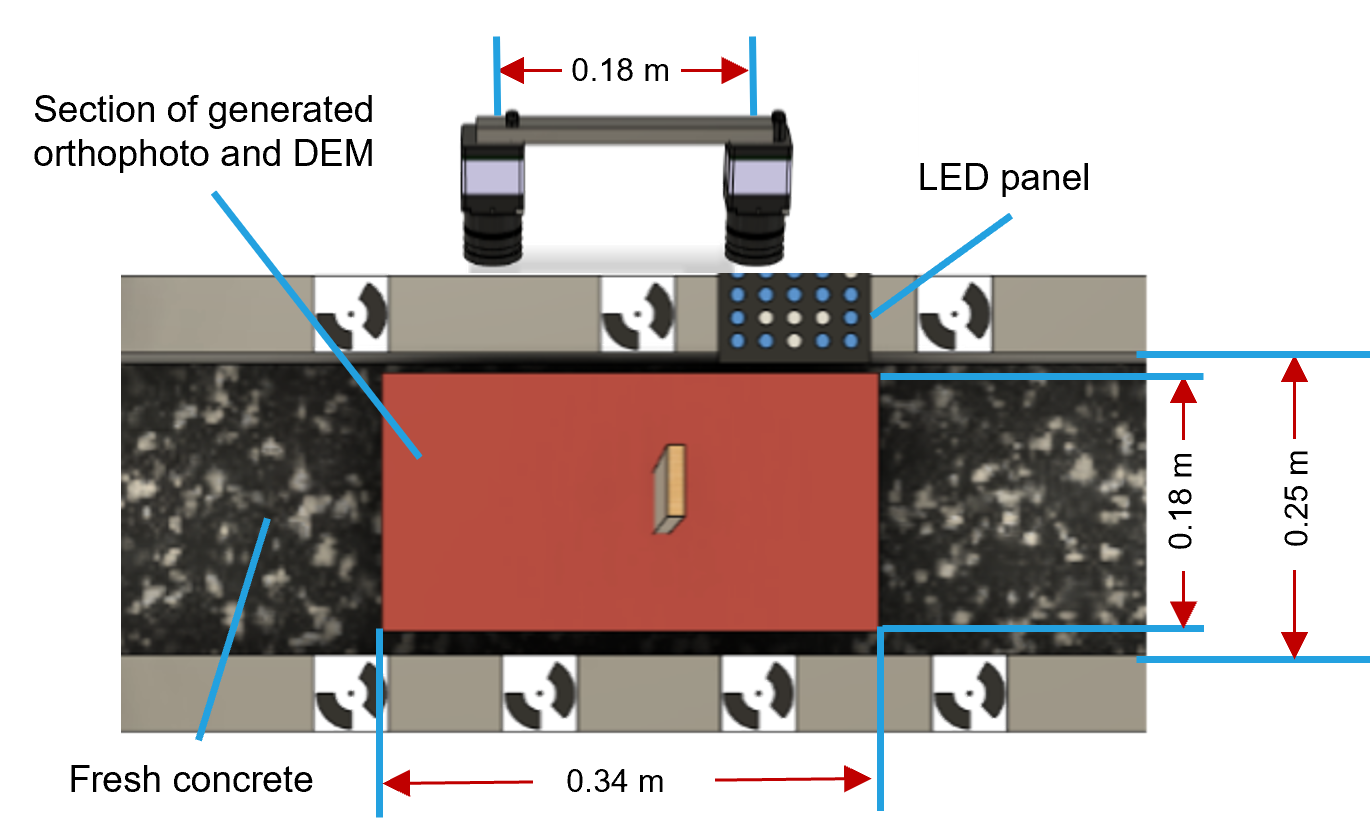}\label{fig:schematic}}
\subfloat[Set up during the experiment.] 
{\includegraphics[width=0.5\textwidth]{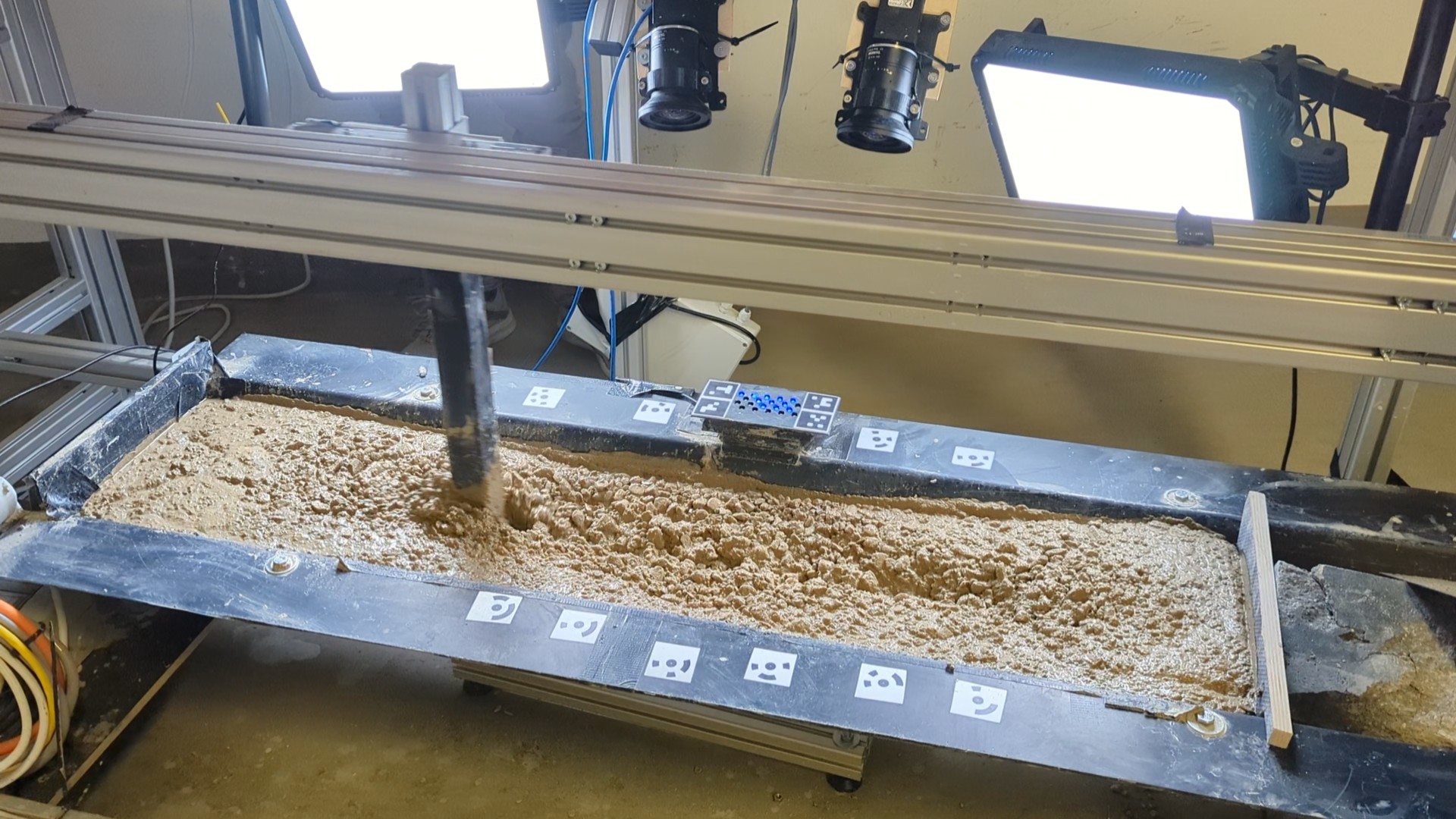}\label{fig:live}}
\caption{Experimental set up for generating the data set.}
\label{fig:setup} 
\end{figure*}
\subsection{Training}\label{sec:Training}
For the optimisation of the network weights $\omega$ the Mean Squared Error (MSE) is used as loss function during training. The weights $\omega$ are iteratively adjusted during training in order to minimize the  resulting loss.
The loss is computed for a mini-batch consisting of $N$ samples, each associated with the state vector $C$ containing the $k=1...K$ target parameters $y^k$, where $K=3$ in this paper. To calculate the loss, the squared differences between the reference values $y^k$ and the predicted values $\hat{y}^k$ are determined and averaged over all parameters and samples in a batch, such that
\begin{equation}
    L_{\text{MSE}}(\omega) = \frac{1}{N \cdot K} \sum_{k=1}^K \sum_{n=1}^N (y^{k}_{n} - \hat{y}^{k}_{n})^2 . 
    \label{eq:lossCNN}
\end{equation}
Weight decay is used during training to reduce over-fitting and to encourage better regularisation. This method adds a penalty for large weights, multiplied by a factor $\lambda$.
This leads to an additional term in the final loss function
\begin{equation}
    L(\omega) = L_\text{MSE}(\omega) + \lambda \cdot \sum \omega^{2} . 
    \label{eq:weightdecayCNN}
\end{equation}
Since the reference values of the slump flow diameter, the yield stress and the plastic viscosity are normalized to the same range the loss of each output does not have to be weighted to ensure an equal influence on the training.

\section{Data generation}\label{data}
\subsection{Image acquisition}
In order to train and test the proposed method we generate an extensive data set: 
Similar to \cite{ponick2022image} we first design a surrogate-mixing system, which consists of a channel, in which fresh concrete is filled, and a paddle shaped mixing tool, which moves through the fresh concrete in direction of the channel. 
The paddle moves on a linear trajectory to simulate the mixing process in an industrial mixer. The fresh concrete is mixed shortly before filling it into the channel. While the paddle is moving, the camera set up (2 grasshopper 3 USB RGB cameras with a focal length of 8 mm) records images of the paddle moving through the concrete. The acquired images have a size of 1920 x 1200 px. In Fig.~\ref{fig:setup}, the schematic set up and the set up during one experiment are shown. In total, 45 different concretes are prepared during the experiments, 5 of which contained recycled aggregates. The mix designs of the concretes all differ due to variations of the water-cement ratio, paste content, grading curve, cement type, sand-lime powder and additive content. Per experiment, we acquire images in 14 so called runs. In each run the paddle moves six times back and forth in the channel, while the cameras record 1300 images. Before each run, the fresh concrete is first briefly mixed manually and then the surface is smoothed. In runs 1-7, the frame rate of the cameras is set to 30 frames per second (fps), and the paddle moves with a velocity of 0.2 $\frac{m}{s}$. In runs 8-14 the frame rate is increased to 60 fps and the velocity of the paddle is set to 0.45 $\frac{m}{s}$. 
\\
\\
In order to maximize the information about the flow behaviour, we only use images which show the concrete directly after the paddle has moved through the material. As a consequence, the paddle is visible in the images. To eliminate effects stemming from the paddle itself we correct $D$ after matching and only use heights below a certain threshold to generate $O$. For generating $O$ and $D$ it is necessary that both images are taken at exactly the same time. To ensure this requirement, a panel with 20 LEDs is installed on the edge of the channel. The LED panel is visible in both images and the LEDs are systematically switched on and off in millisecond intervals. 
Through the changing constellations over time, a time stamp is generated for each image, which is used to verify that both images of an image pair are synchronized.\\ \\
Regarding the fresh concrete properties it is assumed that these remain constant during the short time (approx. 44 and 22 sec) of one run. Therefore, every image pair of a run is associated with the same time stamp, namely that of the central image pair of that run. Consequently, $\Delta_t$ represents the time difference between the associated time stamp of the image pair and the point in time at which the fresh concrete properties are to be predicted.

\subsection{Reference values}
The reference values for the training and testing are measured in parallel to the acquisition of the images. The slump flow diameter $\delta$ is measured with the slump test \citep{flowtabletest} and the yield stress $\tau_0$ and plastic viscosity $\mu$ are measured with a rheometer. For this measurement a \textit{eBT-V} rheometer from \textit{Schleibinger}\footnote{\url{http://www.schleibinger.com/}} is used. The rheological parameters can then be derived from the resulting flow curves. The rheological parameters are determined according to the method presented in \citep{feys2013extension}. To consider the change of the fresh concrete properties over time induced by the chemical hydration of the cement contained in the concretes, the slump test and the rheometer measurement are repeated in intervals of about 30 minutes. The first slump test is always carried out directly after the end of the mixing process of approx. 9 minutes after water addition. Since the slump test and the rheometer measurement are independent from each other, the two measurements have different time stamps. Consequently, $\Delta_t$ consists of two time differences: The first is the time difference between image acquisition and the time at which the slump test is carried out, the second is the time difference between image acquisition and the time at which the rheometer measurement is carried out. The wide range of the reference values and time differences of the data set is shown in Tab.~\ref{tab:reftime}. As some reference measurements are carried out before images are taken, there are also negative values for time differences.
\begin{table}[h!]
	\centering
	\caption{Range of reference values and time differences.}
		\begin{tabular}{|l | c c c c |} \hline	
		& & & & \\		
		  &  Max & Min & Mean & St. deviation\\ \hline \hline
		& & & &\\
		 $\delta$ [cm] & 63.50 & 30.00 & 43.97 & 7.47\\ 
		& & & &\\
		 $\tau_0$ [Pa] & 585.40 & 65.84 & 223.97 & 109.25\\
		& & & &\\
		 $\mu$ [Pa $\cdot$ s] & 121.91 & 19.76 & 49.10 & 17.74 \\
		 & & & &\\
		 $\Delta_t$ [min] & 87.16 & -49.88 & 15.18 & 27.82 \\
		  \hline		
		\end{tabular}	
\label{tab:reftime}
\end{table}  
\section{Experiments}\label{experiments}
In the experiments, the influence of different inputs on the performance of the CNN is investigated. Different input combinations are assembled to train the CNN and the resulting performance metrics are compared.
In particular, the influence of $O$, $D$, $OF$ and $m$ are investigated. Moreover, it is shown, how the accuracy of the predictions changes if the predictions from different input sets for the same reference values are averaged beforehand. At the end, examples of the time-dependent prediction of the behaviour of the slump flow diameter are shown.  
\subsection{Training configuration}
The CNN predicts the three parameters $\delta_{\Delta_t}$, $\tau_{0,\Delta_t}$, and $\mu_{\Delta_t}$ in a multi-task learning manner.
Since two independent measurements (slump test and rheometer measurement) are used to generate the reference values, the inputs $O$, $D$ and $OF$ have to be assigned to one reference combination of $\delta_{\Delta_t}$ respectively $\tau_{0,\Delta_t}$ and $\mu_{\Delta_t}$. 
For each concrete, all possible reference combinations are generated, and each input set ($O_i$, $D_i$ and $OF_{i,i+1}$) is assigned to one reference combination. $m$ always remains the same for the input sets of a concrete, except for the information about paddle velocity and recording frequency of the cameras.
Subsequently, the two time differences of $\Delta_t$ are calculated for each input set based on the assigned reference combination. Sometimes $O_{i+1}$ is missing because the paddle is not visible for several time steps, the end of the run is reached or a framedrop occurs. An input set is only generated if for $O_i$ also $O_{i+1}$ is present, otherwise $OF_{i,i+1}$ could not be calculated. The first 20 input sets of each run are not used, as the paddle had not yet driven far enough through the concrete to significantly change its surface. In total, the data set consists of 313,615 input sets. 
\\
Training is performed with a five-fold cross-validation. The 45 concretes are divided into 5 sets with 9 concretes (i.e. concrete compositions) each. For this purpose, the concretes are first sorted by the length of the first slump flow diameter $\delta_1$, which is determined for each concrete directly after the mixing process. Then, the sorted concretes are divided in three groups, one of each containing the data of the 15 concretes with the largest, the intermediate, and the smallest $\delta_1$, respectively. From each group three concretes are randomly assigned to one of the 5 sets to guarantee a balanced distribution. As there are a total of 5 concretes with recycled aggregate, we make sure that each set has to contain exactly one of these concretes. In each cross-validation step one set is used as the test set. The validation set, containing 5 concretes, is randomly formed by the concretes of the remaining sets, again by taking $\delta_1$ into account and with the condition that it must contain exactly one concrete with recycled aggregate. The remaining 31 concretes form the training set. For two concretes the yield stress and the plastic viscosity are not taken into account, as the corresponding measured reference values are not plausible. For training, only the loss for the slump flow diameter for these concretes is used. For the evaluation, which is explained in the following, the predictions of the yield stress and the plastic viscosity of these concretes are not considered. \\
\\
To train the network, Stochastic Gradient Descent (SGD) with a Nesterov momentum of $\beta=0.99$ based on \citep{sutskever2013importance} is used. The learning rate is set to a value of 5 $\cdot$ 10$^{-3}$, which showed the best results in preliminary experiments. The weight decay parameter is set to $\lambda=1 \cdot 10^{-3}$. The network is trained from scratch and the weights are initialised using the method presented in \cite{he2015delving}, whereby a uniform distribution is used. For training, data augumentation is used. The brightness and contrast of $O$ are each changed with a factor that is randomly determined for each $O$ in each iteration with a uniform distribution in an interval of 0.85 to 1.15 for the brightness and in an interval of 0.75 to 1.25 for the contrast. For the data augumentation of $D$, an offset is determined randomly. For this purpose, a factor is randomly determined with a uniform distribution in an interval from -0.07 to 0.07. The offset is then determined by multiplying the factor by the standard deviation of $D$ in the training data set. The procedure for determining the offsets for the two channels in $OF$ is analogous to that of $D$. To evaluate the results, the mean absolute error
\begin{equation}
    \epsilon_{abs} = \frac{1}{A} \sum^{A}_{a=1} \frac{1}{J} \sum^{J}_{j=1} | y^a_j - \hat{y}^a_j | 
    \label{eq:mae}
\end{equation}
and the mean relative error 
\begin{equation}
    \epsilon_{rel} = \frac{1}{A} \sum^{A}_{a=1}  \frac{1}{J} \sum^{J}_{j=1} \frac{| y^a_j - \hat{y}^a_j |}{y^a_j} \cdot 100  
    \label{eq:mae}
\end{equation}
for a set of $A$ concretes, each with $J$ input sets for each output, are computed. This means that every concrete has the same weight in the evaluation, even if it may have less input sets. After each training epoch, the network is evaluated on the validation set. To determine the best weights, the three $\epsilon_{rel}$-values (one for each output) of the validation set are averaged. The weights with the lowest averaged $\epsilon_{rel}$-value are chosen for testing. Since no significant improvements of the loss are observed after only a few epochs, the number of training epochs is restricted to 5. \\
\\
Finally for each input and reference value the data in the training, validation and test set are normalized to a mean of 0 and a standard deviation of 1.
This is described by 
\begin{equation}
    nd_c = \frac{d_c - m_{t,c}}{std_{t,c}} ,
    \label{eq:norm}
\end{equation}
where $d_c$ represents the data and $nd_c$ the normalized data, each of the data category $c$ (e.g.\ the input $O$ or the reference value $\delta$). $m_{t,c}$ and $std_{t,c}$ represent the mean and standard deviation of the corresponding training set for the data category $c$. The values in $m$ from the mix design are not normalized, as the values are always between 0 and 2. Note, that for the determination of the evaluation metrics the reference values and the corresponding outputs are converted to the original range of reference values. \\

\begin{table*}[ht]
\centering
	\caption{Mean relative and absolute error for different input combinations, whereas $O$ represents orthophoto, $D$ depth elevation map, $OF$ optical flow image and $m$ mix design information (the values in bold show the best performance in the respective category).}
\begin{tabular}{|c c| cccccc|} \hline
& & & & & & &  \\
   & & $O$+$D$+$m$ & $O$+$D$+$m$+$OF$ & $O$+$D$ & $O$+$m$ & $D$+$m$ & $D$+$m$+$OF$ \\ \hline \hline  
    & & & & & & &  \\
\multirow{2}{*}{$\delta_{\Delta_t}$}&$\epsilon_{rel}$ [\%] & 6.87 & 6.94 & 7.07 & 6.97 & 7.00 & \textbf{6.86}         \\ 
& $\epsilon_{abs}$ [cm]& 3.00 & 3.03 & 3.07 & 3.05 & 3.04 & \textbf{2.99} \\ 
& & & & & & &  \\
\multirow{2}{*}{$\tau_{0,\Delta_t}$}&$\epsilon_{rel}$ [\%] &26.49       & 26.08             & 28.29    & 26.68  & 26.07   & \textbf{25.30}        \\ 
& $\epsilon_{abs}$ [Pa] &53.90 & 52.96 & 55.78 & 53.65 & 54.31 & \textbf{52.05} \\
& & & & & & &  \\
\multirow{2}{*}{$\mu_{\Delta_t}$}&$\epsilon_{rel}$ [\%] &26.45       & 26.58             & 29.42    & 27.84  & \textbf{23.99}   & 24.23        \\ 
&$\epsilon_{abs}$ [Pa $\cdot$ s] & 12.12 & 12.40 & 13.28 & 12.70 & \textbf{11.33} & 11.52 \\\hline
\end{tabular}
\label{tab:result_combinations}
\end{table*}
\subsection{Results and discussion}
\subsubsection{Influence of different input combinations:}
To determine the influence of different input combinations the mean relative error $\epsilon_{rel}$ and the mean absolute error $\epsilon_{abs}$ of the test sets are used. For each input combination the above described cross validation is carried out and is repeated for two times. Afterwards the overall mean relative and absolute errors are computed, by taking the mean of the mean relative and absolute errors of all test sets, including the test sets from the repeated cross validations. The results are shown in Tab.~\ref{tab:result_combinations}. In total, six different combinations are investigated. 
\\
\\
It can be seen that the variations of the inputs have only a very limited influence on the accuracy of the predictions of $\delta_{\Delta_t}$. The major differences occur in the accuracy of the predictions of $\tau_{0,\Delta_t}$ and $\mu_{\Delta_t}$. In particular the input $m$ seems to be beneficial for the prediction of these parameters. This can be seen if one compares the results of the combinations $O$+$D$+$m$ and $O$+$D$ (note that in the input combination without $m$ only the information about the used materials are omitted). By comparing the results of combination $O$+$D$+$m$ and $O$+$m$ it can be seen that using the input $D$ also increases the accuracy of the predictions. The input $OF$ has a positive influence on the predictions of $\tau_{0,\Delta_t}$ as can be seen by comparing the results of $O$+$D$+$m$ and $O$+$D$+$m$+$OF$ or $D$+$m$ and $D$+$m$+$OF$. However, the results with and without $O$ are counter-intuitive. The predictions from input combinations with $O$ have a worse accuracy than the predictions from input combinations without $O$. By comparing the results from $O$+$D$+$m$ and $D$+$m$ or $O$+$D$+$m$+$OF$ and $D$+$m$+$OF$ it can be seen that the predictions for $\tau_{0,\Delta_t}$ and especially the predictions for $\mu_{\Delta_t}$ are becoming worse by adding $O$ as input. It is noticeable, that the training runs with $O$ as input have a significantly smaller training loss than training runs without $O$ as input. This indicates that the network is overfitting in training when $O$ is used as input. Consequently, the best overall results are achieved if $D$, $m$ and $OF$ are used as input. To gain a deeper understanding of this behaviour, further investigations will be carried out in the future work. \\
\\
When evaluating the results, it should be noted that the reference measurements are only carried out for batch samples (a small part of the concrete) and the slump flow diameter is determined manually. These circumstances are also reflected in the average precision of the slump test, which is 2.46 cm \citep{flowtabletest}. Furthermore, as all concretes have different mix designs, there is no concrete in the test set that has the same mix design as a concrete in the training or validation set. The results are therefore already within an acceptable range. In general, the prediction of $\delta_{\Delta_t}$ has a much higher accuracy than the predictions for $\tau_{0,\Delta_t}$ and $\mu_{\Delta_t}$. One reason for that can be the much wider range of values and the significantly higher ratio of standard deviation to mean value of $\tau_{0,\Delta_t}$ and $\mu_{\Delta_t}$ (see Tab.~\ref{tab:reftime}).

\subsubsection{Influence of averaging predictions:}
In order to investigate whether the deviations in the predictions are random rather than systematic, predictions from different input sets for the same reference value are averaged beforehand. In Tab.~\ref{tab:result_averaging} it is shown for the example of the input combination $D$+$m$+$OF$ how the accuracy of the predictions changes when they are averaged beforehand. The second column of the table shows the results where the predictions of the same reference combinations within a run are averaged beforehand. On average, these are approx. 40 predictions. The third column shows the results in which all predictions with the same reference combination are averaged over all 14 runs (approx. 540 predictions). The fourth column shows the results where the predictions are averaged for the same slump flow diameter value respectively the same yield stress and plastic viscosity, which is on average about 1900 averaged predictions in each case. It can be seen that the accuracy increases the more predictions are averaged. This indicates that a part of the deviations are indeed random.
\begin{table}[h!]
\centering
	\caption{Influence of averaging multiple predictions of the mean relative and absolute error for the example of the results of the input combination $D$+$m$+$OF$ (the values in bold show the best performance in the respective category; note that the first column is identical to the results of $D$+$m$+$OF$ in Tab.~\ref{tab:result_combinations}).}
\begin{tabular}{|cc|cccc|} \hline
 & & & & & \\ 
 \multicolumn{2}{|c|}{Approx. averaged} &\multirow{2}{*}{1} &\multirow{2}{*}{40} & \multirow{2}{*}{540} & \multirow{2}{*}{1900} \\
 \multicolumn{2}{|c|}{predictions} &          &      &      &    \\ \hline \hline
    & & & & & \\ 
\multirow{2}{*}{$\delta_{\Delta_t}$} & $\epsilon_{rel}$ [\%] & 6.86      & 6.50      & 6.29      & \textbf{6.27} \\
 & $\epsilon_{abs}$ [cm] & 2.99 & 2.84 & 2.75 & \textbf{2.74} \\
 & & & & & \\ 
\multirow{2}{*}{$\tau_{0,\Delta_t}$} & $\epsilon_{rel}$ [\%] & 25.30     & 23.68     & 22.53     & \textbf{22.47} \\
 & $\epsilon_{abs}$ [Pa] & 52.05 & 48.28 & 45.76 & \textbf{45.62}\\ 
  & & & & & \\
\multirow{2}{*}{$\mu_{\Delta_t}$} & $\epsilon_{rel}$ [\%] & 24.23     & 23.53     & 22.87     & \textbf{22.84}   \\
 & $\epsilon_{abs}$ [Pa $\cdot$ s] & 11.52 & 11.17 & 10.82 & \textbf{10.81}\\ \hline
\end{tabular}
\label{tab:result_averaging}
\end{table}

\subsubsection{Time-dependent prediction model for fresh concrete properties:}
\begin{figure}[h!]
\centering
\includegraphics[width=1.0\columnwidth]{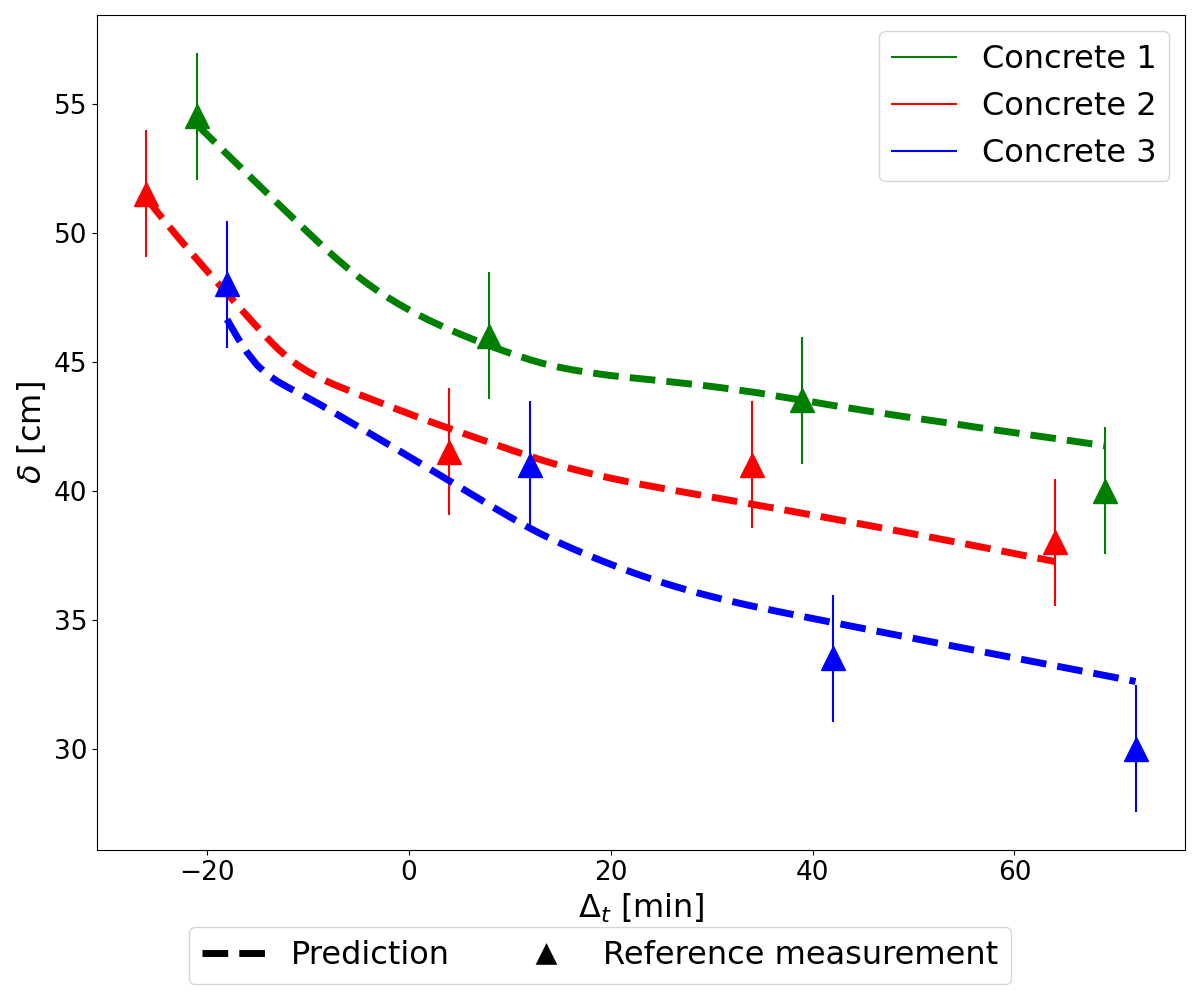}
\caption{Three examples for the prediction of the time-dependent behaviour of the slump flow diameter.}
\label{fig:pred} 
\end{figure}

Since the network receives the time difference between the moment in time at which the images are recorded and the time at which the properties of the fresh concrete are to be predicted, the network implicitly learns how the properties of the concrete change over time. This can be used to not only predict the properties at a certain point in time, but also continuously over the entire fresh concrete age. In Fig.~\ref{fig:pred} examples are shown how such a continuous prediction of the slump flow diameter over a time interval looks like. Note, that at this stage of the research only concretes which exhibit a more or less pronounced decrease in consistency over time were investigated. The model is thus only trained to identify and quantify this specific behaviour, which can be traced back to the type of chemical admixtures used in the project. Changing the admixture as to yield a steady or even an increase in flow over time will be studied in future and will certainly require an adaption of the model or at least its training.\\
\\
To generate the predictions, one of the respective runs with all the input sets it contains is used for each concrete to predict the slump flow diameter at each minute in the time interval. In this example, the runs of concrete 1, 2 and 3 consist of 549, 466 and 525 input sets, which means that each estimate is the result of averaging 549, 466 and 525 predictions, respectively, $D$+$m$+$OF$ is used as input. The point at which $\Delta_t$ is zero is the time at which the images of the runs are recorded (note that as mentioned before, all images from a run are assigned the timestamp of the central image pair). Beside the continuous predictions, the respective reference values are shown for each concrete. The precision of the slump test is shown as an error bar. \\
Considering the continuous prediction of the parameters, it can be seen that the network has learnt that the slump flow diameter decreases particularly sharply in the first minutes after mixing and then decreases more slowly. For the yield stress and plastic viscosity, it has generally learnt that both parameters increase over time, but the nature of the increase can vary from concrete to concrete, and the prediction is not as robust as for the slump flow diameter. The time-dependent behavior that the network has learned for the slump flow diameter is plausible and shows that the network is able to predict this parameter over a longer period of time, even though the prediction is not yet as robust for every concrete as in the examples shown. 

\section{Conclusion and outlook}\label{conclusion}
The results in this paper show that it is possible to predict the fresh concrete properties with an acceptable accuracy based on images of the fresh concrete flow behaviour, supported by the mix design information. In particular, the slump flow diameter can be predicted with a relatively high accuracy. Furthermore, it could be shown that the time-dependent behaviour of the fresh concrete properties can be learned by the network. This makes it possible to predict the properties of the fresh concrete at specific points in time (e.g.\ at the time of placement) or continuously over time already during the mixing process. Furthermore, it has been shown that the use of images and information from the mix design improves the results compared to the use of images alone. However, the information from the mix design in our dataset probably contains only relatively small uncertainties, which is why it has a relatively high positive influence on the result. Such a low level of uncertainty cannot always be assumed. The results have also shown that a significant improvement is possible if predictions of the same values are averaged beforehand.\\
\\
In future work, the counter-intuitive effect of orthophotos on the results will be further investigated. As the results have shown that optical flow images can have a positive impact on the outcome, future work will focus on utilising more of the information contained in image sequences (e.g.\ with transformer-based models). Also, experiments will be conducted with an industrial mixer to test the methodology under realistic conditions. In addition, for an industrial application it is necessary to include the environmental parameters (temperature, humidity, etc.) during the transportation of the concrete to the construction site in the prediction in order to be able to reliably predict the properties up to the time of placement.

\section*{Acknowledgments}
This work is supported by the Federal Ministry of Education and Research of Germany (BMBF) as part of the research project ReCyControl [Project number 0336260A], \url{https://www.recycontrol.uni-hannover.de/en/} and by the LUIS computing cluster funded by the German Research Foundation (DFG) - INST 187/742-1 FUGG.

{
	\begin{spacing}{1.17}
		\normalsize
		\bibliography{Literature} 

\begin{thebibliography}{xx}

\bibitem[Chidiac and Mahmoodzadeh, 2009]{chidiac2009plastic}
Chidiac, S., Mahmoodzadeh, F., 2009.
 Plastic viscosity of fresh concrete--A critical review of predictions methods.
 {\em Cement and Concrete Composites}, 31(8), 535--544.

\bibitem[Coenen et al., 2023]{Coenen2023a}
Coenen, M., Beyer, D., Haist, M., 2023.
 {Granulometry Transformer: Image-based Granulometry of Concrete Aggregate for
  an Automated Concrete Production Control}.
 \emph{Proceedings of the 2023 European Conference on Computing in Construction
  (EC3)}, ~4.

\bibitem[Coenen et al., 2021]{Coenen2021a}
Coenen, M., Schack, T., Beyer, D., Heipke, C., Haist, M., 2021.
 {Semi-Supervised Segmentation of Concrete Aggregate Using Consensus
  Regularisation and Prior Guidance}.
 \emph{ISPRS Annals of the Photogrammetry, Remote Sensing and Spatial
  Information Sciences},  V-2-2021, 83--91.

\bibitem[Coenen et al., 2024]{coenen2024deep}
Coenen, M., Vogel, C., Schack, T., Haist, M., 2024.
 Deep Concrete Flow: Deep learning based characterisation of fresh concrete
  properties from open-channel flow using spatio-temporal flow fields.
 {\em Construction and Building Materials}, 411, 134809.

\bibitem[Ding and An, 2018]{ding2018deep}
Ding, Z., An, X., 2018.
 Deep learning approach for estimating workability of self-compacting concrete
  from mixing image sequences.
 {\em Advances in Materials Science and Engineering}, 2018, 1--16.

\bibitem[{EN 12350-5}, 2019]{flowtabletest}
{EN 12350-5}, 2019.
 {Testing Fresh Concrete - Part 5: Flow Table Test. European Committee for
  Standardization}.

\bibitem[Farneb{\"a}ck, 2003]{farneback2003two}
Farneb{\"a}ck, G., 2003.
 Two-frame motion estimation based on polynomial expansion.
 \emph{Image Analysis: 13th Scandinavian Conference, SCIA 2003 Halmstad,
  Sweden, June 29--July 2, 2003 Proceedings 13}, Springer, 363--370.

\bibitem[Feys et al., 2013]{feys2013extension}
Feys, D., Wallevik, J.~E., Yahia, A., Khayat, K.~H., Wallevik, O.~H., 2013.
 Extension of the Reiner--Riwlin equation to determine modified Bingham
  parameters measured in coaxial cylinders rheometers.
 {\em Materials and structures}, 46, 289--311.

\bibitem[Gao and Yan, 2023]{gao2023numerical}
Gao, X., Yan, H., 2023.
 Numerical detection of concrete slump by fusion of target segmentation and
  image classification network.
 \emph{Journal of Physics: Conference Series},  2562number~1, IOP Publishing,
  012023.

\bibitem[González-Taboada et al., 2018]{Gonzalez2018}
González-Taboada, I., González-Fonteboa, B., Martínez-Abella, F., Roussel,
  N., 2018.
 {Robustness of self-compacting recycled Concrete: Analysis of Sensitivity
  Parameters}.
 {\em Materials and Structures}, 51(8).

\bibitem[Green, 2016]{Green2016}
Green, B., 2016.
 {\em {Productivity in Construction: Creating a Framework for the Industry to
  Thrive}}.
 Chartered Institute of Building (CIOB).

\bibitem[Guo et al., 2022]{guo2022real}
Guo, P., Du, J., Bao, Y., Meng, W., 2022.
 Real-time video recognition for assessing plastic viscosity of
  ultra-high-performance concrete (UHPC).
 {\em Measurement}, 191, 110809.

\bibitem[Haist et al., 2022]{Haist2022}
Haist, M., Heipke, C., Beyer, D., Coenen, M., Vogel, C., Schack, T., Ponick,
  A., Langer, A., 2022.
 {Digitization of the Concrete Production Chain using Computer Vision and
  Artificial Intelligence}.
 \emph{Proceedings of the 6th fib Congress}, 434--443.

\bibitem[He et al., 2015]{he2015delving}
He, K., Zhang, X., Ren, S., Sun, J., 2015.
 Delving deep into rectifiers: Surpassing human-level performance on imagenet
  classification.
 \emph{Proceedings of the IEEE international conference on computer vision},
  1026--1034.

\bibitem[Hosseinzadeh et al., 2023]{hosseinzadeh2023prediction}
Hosseinzadeh, M., Dehestani, M., Hosseinzadeh, A., 2023.
 Prediction of mechanical properties of recycled aggregate fly ash concrete
  employing machine learning algorithms.
 {\em Journal of Building Engineering}, 107006.

\bibitem[IEA, 2022]{IEAcement}
IEA, 2022.
 {Cement, IEA, Paris \url{https://www.iea.org/reports/cement}, License: CC BY
  4.0}.

\bibitem[Kina et al., 2021]{kina2021comparison}
Kina, C., Turk, K., Atalay, E., Donmez, I., Tanyildizi, H., 2021.
 Comparison of extreme learning machine and deep learning model in the
  estimation of the fresh properties of hybrid fiber-reinforced SCC.
 {\em Neural Computing and Applications}, 33, 11641--11659.

\bibitem[Li and An, 2014]{li2014method}
Li, S., An, X., 2014.
 Method for estimating workability of self-compacting concrete using mixing
  process images.
 {\em Computers and Concrete}, 13(6), 781--798.

\bibitem[Lux et al., 2023]{lux2023classification}
Lux, J., Hoong, J. D. L.~H., Mahieux, P.-Y., Turcry, P., 2023.
 Classification and estimation of the mass composition of recycled aggregates
  by deep neural networks.
 {\em Computers in Industry}, 148, 103889.

\bibitem[Navarrete et al., 2023]{navarrete2023predicting}
Navarrete, I., La~F{\'e}-Perdomo, I., Ramos-Grez, J.~A., Lopez, M., 2023.
 Predicting the evolution of static yield stress with time of blended cement
  paste through a machine learning approach.
 {\em Construction and Building Materials}, 371.

\bibitem[Nguyen et al., 2020]{nguyen2020prediction}
Nguyen, T.-D., Tran, T.-H., Hoang, N.-D., 2020.
 Prediction of interface yield stress and plastic viscosity of fresh concrete
  using a hybrid machine learning approach.
 {\em Advanced Engineering Informatics}, 44.

\bibitem[Ponick et al., 2022]{ponick2022image}
Ponick, A., Langer, A., Beyer, D., Coenen, M., Haist, M., Heipke, C., 2022.
 Image-Based Deep Learning for Rheology Determination of Bingham Fluids.
 {\em The International Archives of the Photogrammetry, Remote Sensing and
  Spatial Information Sciences}, 43, 711--720.

\bibitem[Schack et al., 2023a]{schack2023bildbasierte1}
Schack, T., Coenen, M., Haist, M., 2023a.
 Bildbasierte Frischbetonprüfung – Teil 1: Konsistenz und Leimgehalt des
  Frischbetons.
 {\em Beton- und Stahlbetonbau}, 118(4), 220-228.

\bibitem[Schack et al., 2023b]{schack2023bildbasierte2}
Schack, T., Coenen, M., Haist, M., 2023b.
 Bildbasierte Frischbetonprüfung – Teil 2: Granulometrische Eigenschaften
  der Gesteinskörnung.
 {\em Beton- und Stahlbetonbau}, 118(8), 556-564.

\bibitem[Schack et al., 2023c]{schack2023bildbasierte3}
Schack, T., Coenen, M., Haist, M., 2023c.
 Bildbasierte Frischbetonpr{\"u}fung: Teil 3: Homogenit{\"a}t des Frischbetons.
 {\em Beton- und Stahlbetonbau}, 118(10), 716-724.

\bibitem[Song et al., 2020]{song2020deep}
Song, Y., Huang, Z., Shen, C., Shi, H., Lange, D.~A., 2020.
 Deep learning-based automated image segmentation for concrete petrographic
  analysis.
 {\em Cement and Concrete Research}, 135, 106118.

\bibitem[Sutskever et al., 2013]{sutskever2013importance}
Sutskever, I., Martens, J., Dahl, G., Hinton, G., 2013.
 On the importance of initialization and momentum in deep learning.
 \emph{International conference on machine learning}, PMLR, 1139--1147.

\bibitem[Tuan et al., 2021]{tuan2021situ}
Tuan, N.~M., Van~Hau, Q., Chin, S., Park, S., 2021.
 In-situ concrete slump test incorporating deep learning and stereo vision.
 {\em Automation in Construction}, 121, 103432.

\bibitem[Wallevik, 2006]{wallevik2006relationship}
Wallevik, J.~E., 2006.
 Relationship between the Bingham parameters and slump.
 {\em Cement and concrete research}, 36(7), 1214--1221.

\bibitem[Yahia et al., 2016]{yahia2016concrete}
Yahia, A., Mantellato, S., Flatt, R.~J., 2016.
 Concrete rheology: A basis for understanding chemical admixtures.
 \emph{Science and Technology of Concrete Admixtures}, Elsevier, 97--127.

\bibitem[Yang et al., 2021]{yang2021estimating}
Yang, L., An, X., Du, S., 2021.
 Estimating workability of concrete with different strength grades based on
  deep learning.
 {\em Measurement}, 186.

\bibitem[Yoon et al., 2023]{yoon2023framework}
Yoon, J., Kim, H., Ju, S., Li, Z., Pyo, S., 2023.
 Framework for rapid characterization of fresh properties of cementitious
  materials using point cloud and machine learning.
 {\em Construction and Building Materials}, 400, 132647.

\bibitem[Zhang et al., 2022]{zhang2022predicting}
Zhang, X., Akber, M.~Z., Zheng, W., 2022.
 Predicting the slump of industrially produced concrete using machine learning:
  A multiclass classification approach.
 {\em Journal of Building Engineering}, 58, 104997.

\end{thebibliography}
	\end{spacing}
}

\end{document}